\documentclass[letterpaper, 10 pt, conference]{ieeeconf}

\usepackage{microtype}
\usepackage{subcaption}
\usepackage{booktabs} 
\usepackage{wrapfig}
\usepackage{tablefootnote}
\usepackage{xspace}
\usepackage{xcolor}
\usepackage{hyperref}
\usepackage{comment}
\usepackage[linesnumbered, ruled]{algorithm2e}

\usepackage{amsmath}
\usepackage{amssymb}
\usepackage{mathtools}
\usepackage{amsthm}

\usepackage{bbm}

\usepackage{amsmath,amssymb,amsthm}

\usepackage[capitalize,noabbrev]{cleveref}

\usepackage{caption}
\usepackage{subcaption}
\usepackage{multirow}
\usepackage{duckuments}
\usepackage{tikz}
\usepackage{pifont}
\usepackage{float}

\usepackage[utf8]{inputenc} 
\usepackage[T1]{fontenc}    
\usepackage{hyperref}       
\usepackage{url}            
\usepackage{booktabs}       
\usepackage{amsfonts}       
\usepackage{nicefrac}       
\usepackage{microtype}      
\usepackage{xcolor}         

\usepackage{amsmath,amsfonts}
\usepackage{array}
\usepackage{enumerate}
\usepackage[utf8]{inputenc}
\usepackage[T1]{fontenc}
\usepackage{amsfonts}
\usepackage{amssymb}
\IEEEoverridecommandlockouts 
\usepackage{breqn}
\usepackage{textcomp}
\usepackage{stfloats}
\usepackage{url}
\usepackage{verbatim}
\usepackage{graphicx}
\usepackage{cite}
\usepackage{lipsum}
\usepackage{mathtools}
\usepackage[linesnumbered, ruled]{algorithm2e}

\usepackage{hyperref}
\hypersetup{
colorlinks=true,
linkcolor=blue,
filecolor=magenta,
urlcolor=blue,
}

\usepackage[font=footnotesize]{caption}

\title{\LARGE \bf

Human-Robot Cooperative Distribution Coupling for Hamiltonian-Constrained Social Navigation
}

\author{Weizheng Wang$^{1}$, Chao Yu$^{2}$, Yu Wang$^{2}$, and Byung-Cheol Min$^{1}$ 
\thanks{$^{1}$SMART Laboratory, Department of Computer and Information Technology, Purdue University, West Lafayette, IN, USA. {\tt\small{[wang5716, minb]@purdue.edu}.}
 }
\thanks{$^{2}$Department of Electronic Engineering, Tsinghua University, Beijing, China. {\tt\small{[yuchao, yu-wang]@mail.tsinghua.edu.cn}.}
}}

\begin{document}

\maketitle

\begin{abstract}

Navigating in human-filled public spaces is a critical challenge for deploying autonomous robots in real-world environments. This paper introduces NaviDIFF, a novel Hamiltonian-constrained socially-aware navigation framework designed to address the complexities of human-robot interaction and socially-aware path planning. NaviDIFF integrates a port-Hamiltonian framework to model dynamic physical interactions and a diffusion model to manage uncertainty in human-robot cooperation. The framework leverages a spatial-temporal transformer to capture social and temporal dependencies, enabling more accurate spatial-temporal environmental dynamics understanding and port-Hamiltonian physical interactive process construction. Additionally, reinforcement learning from human feedback is employed to fine-tune robot policies, ensuring adaptation to human preferences and social norms. Extensive experiments demonstrate that NaviDIFF outperforms state-of-the-art methods in social navigation tasks, offering improved stability, efficiency, and adaptability\footnote{The experimental videos and additional information about this work can be found at: \url{https://sites.google.com/view/NaviDIFF}.}.

\label{web}

\end{abstract}

\section{Introduction}
\begin{figure}[!t]
\centering
\includegraphics[width=0.90\columnwidth]{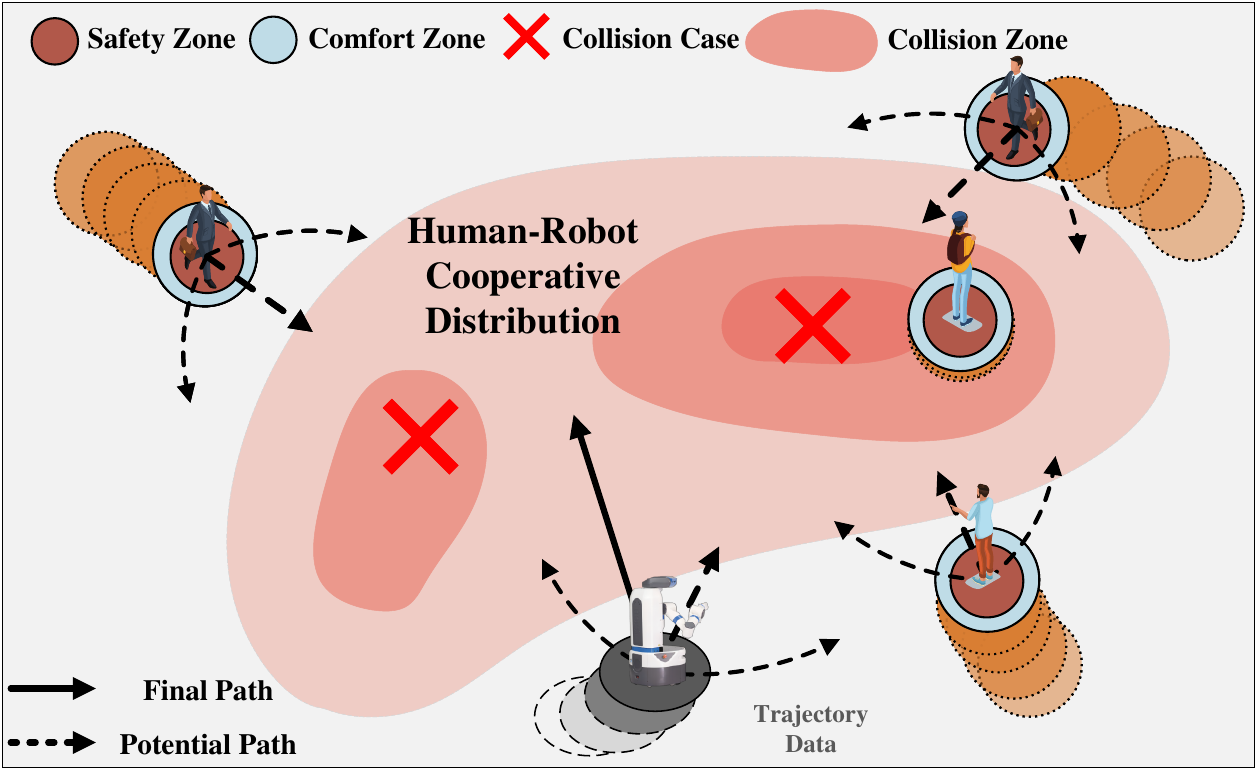}
\vspace{-4pt}
\caption{An illustration of a social navigation task: the mobile robot must navigate through pedestrians while maintaining appropriate social distances and finding a feasible path.}
\vspace{-20pt}
\label{fig:F1}
\end{figure}

Navigating in human-filled public spaces is a fundamental yet complex task for robotics and is critical for the successful deployment of autonomous robots in real-world environments \cite{khatib1986real, ido2009indoor, pereira2009robot, newman2009navigating, biswas2013localization, kretzschmar2016socially, fan2020distributed, devo2020towards, yao2021singularity, legrobot2023TRO}. As illustrated in Fig.~\ref{fig:F1}, robots not only need to avoid collisions while reaching their destinations, but they must also adhere to social norms, requiring sophisticated human-robot interaction (HRI) and socially-aware path planning.

Recent research in social navigation has made notable progress by integrating techniques from deep reinforcement learning (DRL) \cite{wang2023navistar, wang2024multi}, analytical mechanics \cite{mavrogiannis2021hamiltonian, 2023Hamilton}, and sociology \cite{trautman2015robot, mavrogiannis2019multi, mavrogiannis2022social, samavi2024sicnav}. However, current approaches, whether implicitly coupled \cite{wang2023navistar, wang2024multi, liu2023dsrnn2, SunM-RSS-21} or decoupled \cite{cao2019dynamic, du2011robot, thompson2009probabilistic, vasquez2014inverse, kim2016socially, knepper2012pedestrian}, still face two primary  challenges: uncertainty in understanding human-robot cooperation and the lack of models that can fully capture the physical dynamics of HRI.

Many methods that couple HRI with path planning struggle to account for the physical interactions between humans and robots due to simplified environmental dynamics. This often results in errors in intent inference and fails to capture internal system dependencies. For instance, some approaches assume uniform motion in environmental kinematics \cite{liu2023dsrnn2, CADRL, dsrnn1}, while others introduce non-uniform motion through acceleration terms \cite{wang2024multi}. However, both approaches overlook the importance of maintaining energy balance within the system. While DRL has shown significant potential in robotics \cite{Xu2024UAVRL, yu2022learning, ibarz2021train, kalashnikov2022scaling, xu2020prediction, haarnoja2024learning, rudin2022learning, raffin2022smooth, dalal2021accelerating}, many DRL-based navigation planners rely on simplified models that neglect the physics of human-robot interactions.

To address the limitations in human-robot cooperation and dynamic physical interactions, we introduce a two-pronged approach that combines a port-Hamiltonian (PH) framework with a diffusion model, trained via a DRL algorithm. This approach addresses both the uncertainty of human-robot cooperation and the challenges posed by dynamic physical interactions. First, the PH framework \cite{van2006portintro} models the physical configurations and interactions between humans and robots. Widely applied in robotic control \cite{angerer2017port, 2017hamilarm, altawaitan2024hamiltonian}, multi-agent path planning \cite{sebastian2023lemurs, song2024port}, and neural network systems \cite{massaroli2019port, desai2021port}, PH allows us to capture energy-conserving, closed-loop systems that accurately describe the physical aspects of human-robot interactions, which are crucial for social navigation tasks.

Second, to manage the uncertainty of human-robot cooperation and capture temporal dependencies, we employ a diffusion model combined with a spatial-temporal (ST) transformer. The diffusion neural network \cite{mao2023leapfrog} stochastically infers the distribution of human-robot cooperation styles, addressing uncertainty in intent inference. Meanwhile, the ST transformer \cite{wang2023navistar, wang2024multi} captures both social and temporal dependencies in human behavior, enabling more accurate construction of PH mechanics and predictions of human trajectories.

By combining the PH framework and the diffusion model, we generate robot policies that balance physical interactions with cooperative uncertainties. The Hamiltonian-constrained features of the system are parameterized to generate robot actions using a DRL policy, allowing the robot to adapt its behavior in complex environments. Additionally, reinforcement learning from human feedback (RLHF) \cite{wang2023navistar} is employed to fine-tune robot policies, embedding human preferences and social norms into the navigation strategy.

This paper introduces NaviDIFF, a novel Hamiltonian-constrained socially-aware navigation (HCSAN) framework, with the following key contributions:
\begin{itemize}

\item NaviDIFF is presented as an interpretable social navigation framework that ensures closed-loop stability and models environmental physical dynamics using the PH mechanics. It effectively manages diverse human-robot cooperative distributions through a diffusion neural network and captures social and temporal dependencies via a spatial-temporal (ST) transformer.

\item The framework bridges the gap between implicit and explicit HRI inference and planning by combining Hamiltonian mechanics with DRL. Additionally, RLHF is utilized to fine-tune robot policies based on human feedback, enabling mobile robots to adapt to human preferences and social norms.

\item Extensive experiments are conducted to evaluate the framework’s performance, demonstrating that NaviDIFF outperforms state-of-the-art methods.

\end{itemize}

\section{Related Works}
\noindent\textbf{Port-Hamilton Mechanics:} 
The Hamiltonian approach, rooted in analytical mechanics, evolved from the principle of least action through the Euler-Lagrange equations and the Hamiltonian equations of motion. In contrast, the network approach, originating from electrical engineering, forms the backbone of systems theory. Port-Hamiltonian systems unify these perspectives by linking the network model’s interconnection structure (often referred to as the "generalized junction structure" in bond graph terminology) with a geometric framework defined by a Poisson or Dirac structure. This framework allows Hamiltonian dynamics to model total stored energy, energy dissipation, and system ports.

Neural networks are commonly used for learning control policies, but they often fail to account for energy conservation and kinematic constraints in physical systems, potentially leading to instability. Physics-informed machine learning, which integrates physical laws into the learning process, addresses this limitation and has been applied in centralized controllers for robotics and multi-agent systems \cite{sebastian2023lemurs, angerer2017port, altawaitan2024hamiltonian}. These principles have been extended to cooperative distributed control using port-Hamiltonian neural networks, which model complex systems by combining data from simpler subsystems \cite{neary2023compositional, massaroli2019port}.

While Hamiltonian mechanics have traditionally been applied to centralized control and distributed systems with fixed-time topologies \cite{shi2020neural}, our work introduces a novel approach by modeling human-robot interactions as Hamiltonian terms in a port-Hamiltonian system. This method explicitly captures energy-based dependencies in human-robot interactions, enhancing the modeling of cooperative distributed control in dynamic environments.

\noindent\textbf{Socially Aware Robot Navigation:} 
Advancements in robotics and artificial intelligence \cite{ppo} have led to the development of numerous socially-aware robot navigation planners \cite{wang2023navistar, wang2024multi, SunM-RSS-21}, evolving from early systems like MINERVA \cite{thrun1999minerva} and RHINO \cite{dieterfox1998map}. These planners can be categorized into decoupled and coupled strategies for HRI inference and path planning \cite{cmu}. Decoupled methods \cite{cao2019dynamic, du2011robot, thompson2009probabilistic, vasquez2014inverse, kim2016socially, knepper2012pedestrian} often suffer from issues such as the "freezing robot" problem \cite{trautman2015robot} and "reciprocal dance" \cite{dance} in dynamic environments due to the gap between map scanning and path planning.

Recently, coupled approaches leveraging game theory \cite{turnwald2019human}, geometry analysis \cite{mavrogiannis2019multi}, and variational methods \cite{SunM-RSS-21} have been introduced to address these issues. RL-based planners have also achieved significant success by approximating human-like navigation through environmental dynamics representation. For instance, \cite{CADRL} addresses cooperative collision avoidance via pairwise agent interactions, while \cite{sarl, rgl, everett2018motion} handle crowd-wise interactions. Other methods \cite{liu2023dsrnn2, dsrnn1} use ST graphs to explicitly model HRI, and transformer-based approaches \cite{wang2023navistar} further improve performance.

Recent advances have incorporated human preferences and social norms into robot policies through preference learning \cite{wang2023navistar, SunM-RSS-21} and explored multi-robot learning-based paradigms \cite{wang2024multi}. However, these approaches still fail to fully address the uncertainty in human-robot cooperation and dynamic physical interactions. To overcome these limitations, we propose a physics-informed mobile robot navigation framework that captures energy-based environmental dynamics with a closed-loop stability guarantee. A diffusion neural network is employed to represent HRI cooperative distributions, trained through RLHF to embed social norms.

\section{Preliminary}
\label{sec:Preliminary}

Inspired by recent applications of the Hamiltonian mechanics in robotics \cite{massaroli2019port, desai2021port}, the socially-aware robot navigation (SAN) task can also be modeled as a Hamiltonian-based multi-agent system \cite{sebastian2023lemurs}, meeting the necessary physical criteria. In this context, we formulate the Hamiltonian-constrained socially-aware robot navigation task using the PH mechanics. The PH system offers an interpretable, energy-based model of physical environmental dynamics, ensuring closed-loop energy stability and capturing human-robot interaction patterns. The typical input-state-output port-Hamiltonian system can be defined as follows:
\begin{equation}
\left\{
\begin{aligned}
    \dot{x} &= [\mathbf{J}(x) - \mathbf{R}(x)] ~ \nabla_{x} \mathbf{H}(x) + \mathbf{G}(x)u;\\
    {y} &= \mathbf{G}^{\top}(x) ~ \nabla_{x} \mathbf{H}(x)
\end{aligned}
\right.
\end{equation}
\noindent where $\nabla_{x} := \frac{\partial}{\partial x}$ denotes the gradient function, $\mathbf{H}(x)$ represents the system's total energy,  $\mathbf{J}(x) = - \mathbf{J}(x)^{\top}$ represents the system's interactive energy, $\mathbf{R}(x) = \mathbf{R}(x)^{\top}$ denotes the system's dissipative energy, and $\mathbf{G}(x)$ represents the external input. Here, $u$ is the robot policy input, $x$ is the robot's state, and $y$ represents the conjugated variables of $u$. The PH system can be viewed as an input-state-output framework, where the input and output space are $m$-dimensional vector spaces, and the state space is an $n$-dimensional manifold. Additionally, a ST-graph $\mathcal{G}=(\mathcal{V}, \mathcal{E})$ is constructed using a spatial-temporal transformer network to model the interactions within the multi-agent system for the Hamiltonian mechanics. In this graph, each vertex represents an individual agent, and the spatial or temporal dependencies between them are abstracted as edges.

\begin{figure*}[t]
\centering
     \includegraphics[width=0.94\linewidth]{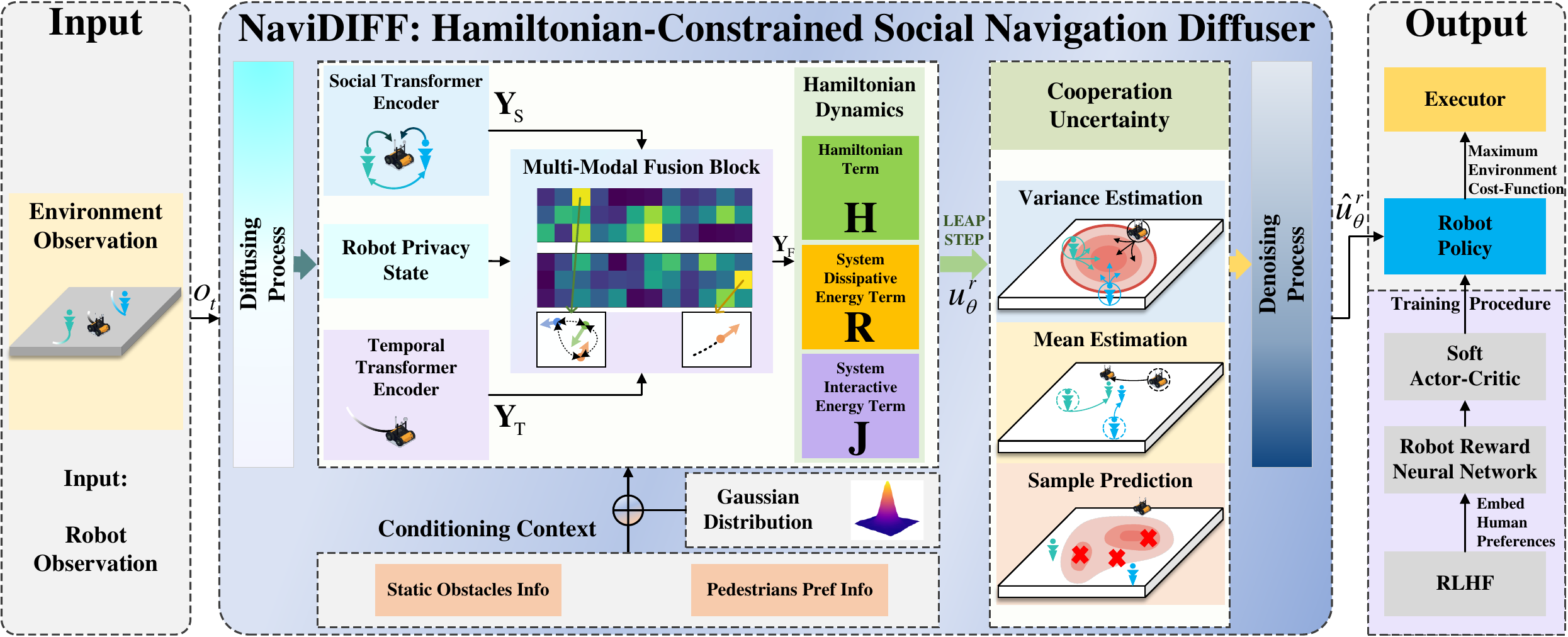}
     \caption{\small NaviDIFF Architecture: NaviDIFF leverages a spatial-temporal transformer to represent Hamiltonian terms, capturing HRI environmental dynamics and ensuring closed-loop stability. It also addresses cooperative uncertainty using a diffusion neural network. Additionally, the framework incorporates reinforcement learning from human feedback (RLHF) and utilizes a large language model (LLM) for enhanced performance.}
          \label{fig:frame}
     \vspace{-15pt}
\end{figure*}

However, the computational complexity of traditional PH solutions \cite{sebastian2023lemurs} poses challenges due to the reliance on a set of partial differential equations (PDE) or ordinary differential equations (ODE). To mitigate this issue, a reinforcement learning (RL) solver is employed to relax the ODE or PDE problem into the stochastic optimization of non-linear systems. 

Formally, the non-communication multi-agent cooperative collision avoidance problem \cite{wang2023navistar, wang2024multi} in the SAN task is modeled as a decentralized particularly observable Markov decision-making process (Dec-POMDP) with the tuple $\langle \mathcal{S}, \mathcal{A}, \mathcal{O}, \mathcal{R}, \mathcal{S}_{0}, \mathcal{P}, \gamma, N \rangle$. Here, $s = [s^{r}, s^{h}] \in \mathcal{S}$ represents the full state of the environment, consisting of the robot state and human state, while $\mathbf{a} =[v_x, v_y] \in \mathcal{A}$ denotes agent velocity actions. The observation $o = [s^{r}, s^{ho}] \in \mathcal{O}$ includes the robot state and observable human state. $\mathcal{R}$ represents the reward function, $\mathcal{S}_{0}$ is the initial state distribution, and $\mathcal{P}(s_{t_{\alpha+1}}|s_{t_\alpha},a_{t_\alpha})$ is the environment transition probability, modeled as a Gaussian distribution with mean $\mu = {s}_{t_\alpha} + T \cdot f_{PH}(s_{t_\alpha},a_{t_\alpha}) $. $\gamma$ is the discount factor, and $N$ represents the number of agents.

The continuous procedure of the PH mechanics is adapted to the discretization of the DRL paradigm over a set of discrete time sequences ${\{t_{\alpha}\}}^{\infty}_{\alpha=0}$, where $(t_{\alpha+1} - t_{\alpha}) = T > 0$. The actions follow the zero-order hold input assumption $\{\forall t \in [t_{\alpha},t_{\alpha+1})~;~ a_t = u_{t_{\alpha}}\}$. We define the PH dependence $\dot{x}$ using the PH function $f_{PH}(\cdot)$, which is represented by a spatial-temporal transformer. The discretization of the continuous environmental dynamics is then abstracted into discrete states and actions, $s_{t} = x_{t_{\alpha}}, a_{t} = u_{t_{\alpha}}$, as follows: 
\begin{equation}
\label{equation:gap}
\begin{aligned}
    \dot{x_t} &= f_{PH}(x_t,u_t), \\
    {s}_{t+1} &= {s}_{t} + T \cdot f_{PH}(s_t,a_t). 
\end{aligned}
\end{equation}

\section{Methodology}
\subsection{Hamiltonian-Constrained Socially Aware Navigation}
In this work, we propose the Hamiltonian-constrained socially-aware navigation (HCSAN) framework for deploying mobile robots in human-shared spaces.  As shown in Fig.~\ref{fig:frame}, NaviDIFF captures HRI spatial-temporal features to explicitly models energy-based environmental dynamics while ensuring closed-loop stability. Subsequently, the physic-informed action execution is diffused to address human-robot cooperative uncertainty. To optimize the performance of social navigation planners, we assume that robots are invisible to pedestrians, and pedestrians are driven by individual behavior policies. Formally, the mobile robot in HCSAN is modeled as an open-loop port-Hamiltonian system, as defined below:
\begin{equation}
\begin{aligned}
    &\dot{x} = f_{PH}(x,u) = [\mathbf{J}(x) - \mathbf{R}(x)] ~ \nabla_{x} \mathbf{H}(x) + \mathbf{G}(x)u.\\
\end{aligned}
\label{eq:4}
\end{equation}
The PH system of the mobile robot adheres to the energy conservation principle, as represented by the power balance equation \cite{ortega2008control}, which satisfies $(\mathbf{J}(x) - \mathbf{R}(x)) + (\mathbf{J}(x) - \mathbf{R}(x))^{\top} = -2 \mathbf{R}(x) \leq 0$:
\begin{equation}
\begin{aligned}
    \dot{\mathbf{H}}(x) &= (\nabla_x \mathbf{H}(x))^{\top} \dot{x} \\ &= -(\nabla_x \mathbf{H}(x))^{\top} \mathbf{R}(x) \nabla_x \mathbf{H}(x) + u^{\top} y ~ \leq u^{\top} y.\\
\end{aligned}
\label{eq:5}
\end{equation}

The SAN PH system, which satisfies Eq.~(\ref{eq:5}), can be considered a passive system. Therefore, by employing the energy-balancing passivity-based control approach \cite{ebpbc2, ebpbc1}, along with energy shaping and damping injection \cite{RL_Hamilton}, we can derive the desired robot policy $u_d$. This achieves the closed-loop dynamics of the SAN system as follows:
\begin{equation}\label{eq:closed_loop_intro}
        \dot{x} = \left( \mathbf{J}_{d}(\mathbf x)  -  {\mathbf{R}}_{d}(\mathbf x)\right) \nabla_{x} \mathbf{H}_{d}(x).
\end{equation}

By aligning Eq.~(\ref{eq:4}) and Eq.~(\ref{eq:closed_loop_intro}), the desired policy for the mobile robot can be defined as:
\begin{equation}
\begin{aligned}
    {u}(x) =& ~\mathbf{G}^{\dagger}(x)(\mathbf J_{d}(x) - \mathbf R_{d}(x)) \nabla_{x} \mathbf{H}_{d}(x) - \mathbf D(x)y\\
    =& ~\mathbf G^{\dagger}(x)((\mathbf J_{d}(x) - \mathbf R_{d}(x)) \nabla_{x} \mathbf{H}_{d}(x) - \\ ~~~~~& (\mathbf J(x) - \mathbf R(x) ) \nabla_{x} \mathbf H(x) )
\end{aligned}
\end{equation}
where $\mathbf{G}^{\dagger}(\mathbf{x}) = \left(\bf G^{\top}(x)\bf G(x)\right)^{-1}\bf G^{\top}(x)$ is the pseudo-inverse of $\mathbf{G}(x)$. $\mathbf D(x)$ denotes the parameter of the damping injection matrix, defined as $\mathbf D(x) = (\mathbf{G}^{\top}(x) \mathbf{G}(x) )^{-1} (\mathbf{J}(x) - \mathbf{R}(x))$.

Eventually, the desired policy of mobile robot $u^{r}_{\theta}$, driven by the PH system and accounting for the underlying energy-based Hamiltonian system dynamics, can be formulated as:
\begin{equation}
\label{equation:PH-result}
\begin{aligned}
u^{r}_{\theta}(x) &= \mathbf{G}^{\dagger}(x)
(\sum_{n \in \mathcal{V}}\left([\mathbf{J}_{ \theta}( x)]_{rn} - [\mathbf{R}_{ \theta}(x)]_{rn}\right){ \nabla_{x_n} \mathbf{H}_{ \theta}( x)} \\ & - (\mathbf{J}( x) - \mathbf{R}(x) ) {\nabla_{x_r} \mathbf{H}(x)} )
\end{aligned}
\end{equation}
where $[\mathbf{J}_{\theta}(x)]_{rn}, [\mathbf{R}_{\theta}(x)]_{rn}$ represents the pairwise energy interaction and energy dissipation between the robot and the $n$-th pedestrian, respectively.

Additionally, if the term $\mathbf G(x)$ is full-rank, the transformation of the SAN Hamiltonian mechanics  from open-loop to close-loop dynamics is exact with respect to the policy input. However, for other agnostic systems, this transformation may not always be successfully  addressed \cite{blankenstein2002matching}. In this context, NaviDIFF not only optimizes the energy-balance PH policy using DRL, rather complex PDE or ODE inference, but also leverages a diffusion neural network to approximate the uncertainty in energy-based HRI dynamics.

\subsection{HRI State Representation and Cooperation Uncertainty}

We extend the hybrid spatial-temporal transformer network from \cite{wang2023navistar} to infer latent spatial-temporal HRI dependencies for Hamiltonian term representation. The HRI spatial attentions $\mathbf{{Y}}_{\rm{S}} \in \mathbb{R}^{N \times T \times d}$ are generated by a spatial transformer to capture human-robot spatial correlations and interactions, while the agent temporal attentions $\mathbf{{Y}}_{\rm{T}} \in \mathbb{R}^{N \times T \times d}$ are captured by a temporal transformer, representing agents' temporal predictability. Once the spatial and temporal HRI features are obtained, they are fused into a latent state $\mathbf{{Y}_{\rm{F}}} \in \mathbb{R}^{N \times T \times d}$ via multi-modal fusion transformer. 

To generate Hamiltonian terms, ${\bf Y}_{{\rm F}}^{\rm}$ is first reshaped into a fused spatial-temporal feature $\mathbf{F}_{\rm R}^{\rm}$, where $\mathbf{F}_{\rm R}^{\rm} = \mathsf{Reshape}({\bf Y}_{{\rm F}}^{\rm})$. The resulting spatial-temporal features are then encoded into energy-based Hamiltonian terms ($\mathbf{R}, \mathbf{J}, \mathbf{H}$) as follows:
\begin{equation} \label{eq:Restimate}
\begin{aligned}
[\mathbf{R}_{ \theta}(\bf x)]_{ij} &= f_{fc}^{R}[-(\mathbf{F}_{\rm R}^{\rm ij}+\mathbf{F}_{\rm R}^{\rm ji})], \quad  \forall j \in [1, N];\\
[\mathbf{R}_{ \theta}(\bf x)]_{ii} &= f_{fc}^{R} [ \mathbf{F}_{\rm R}^{\rm ii} +  {\sum}_{j \in [1,N]}   (\mathbf{F}_{\rm R}^{\rm ij} + \mathbf{F}_{\rm R}^{\rm ji})];    \\
[\mathbf{J}_{ \theta}(\bf x)]_{ij} &= f_{fc}^{J} [{\mathbf{F}_{\rm K}^{\rm ij}}-{\mathbf{F}_{\rm M}^{\rm ji}}], \quad \forall j \in [1,N];  \\
\mathbf{H}_{\theta}(x) &= \mathbf{E}(x) + \mathbf U(x) = f_{fc}^{E}({\bf Y}_{{\rm F}}^{\rm}|o) + f_{fc}^{U}({\bf Y}_{{\rm F}}^{\rm}|o)
\end{aligned}
\end{equation}
\noindent where $f_{fc}^{R}, f_{fc}^{J}, f_{fc}^{E}, f_{fc}^{U}$ are fully connected layers with individual parameters. The system's kinetic energy $\mathbf{E}(x)$ and potential energy $\mathbf U(x)$ are incorporated into the Hamiltonian $\mathbf{H}_{\theta}(x)$.

Here, the system's interactive energy matrix $\mathbf{J}_{\theta}(x)$ represents the physically-informed human-robot energy interaction or exchange, while system's dissipative energy matrix $\mathbf{R}_{\theta}(x)$ captures the negative influence between agents The total energy of the social human-robot system is represented by the Hamiltonian term $\mathbf{H}_{\theta}(x)$. Once these three Hamiltonian components are generated based on the spatial-temporal feature representation of environmental dynamics from the spatial-temporal transformer, the PH-based robot policy is calculated using Eq.~(\ref{equation:PH-result}). 

To further explore the diversity of human-robot cooperation or competition correlations, the cooperation uncertainty of the human-robot system is estimated using a leaping diffusion neural network \cite{mao2023leapfrog}. First, the PH-based robot execution $u^{r}_{\theta}$ is mapped into the human-robot cooperation distribution space through a forward noising process $q(\cdot|\cdot)$. This process begins from the initial step $u^{0}_{\theta}$ and proceed to the $K$-th step $u^{K}_{\theta}$, defined as follows:
\begin{equation}
\begin{aligned}
    q(u^{K}_{\theta}|u^{0}_{\theta}) & = \mathcal{N}(u^{K}_{\theta}; ~ \sqrt{ \textstyle{\prod_{k=1}^{K}}(1-\alpha_k)}u^{0}_{\theta},\\ & ~ (1- \textstyle{\prod_{k=1}^{K}}(1-\alpha_k)) \mathbf{I})
\end{aligned}
\end{equation}
\noindent where $\alpha$ is the noise parameter, and the signal $u^{r}_{\theta}$ is diffused into a Gaussian distribution during the process.

Afterward, NaviDIFF initializes the HRI patterns through an intermediate denoising step $\kappa$, allowing it to skip part of the standard reverse process steps $(K,K-1,\cdots, \kappa)$, as follows:
\begin{equation}
\begin{aligned}
    \mu_{\theta} &= f_{\mu}(u^{r}_{\theta}) ; ~
    \sigma_{\theta} = f_{\sigma}(u^{r}_{\theta}); ~ \mathbb{C} = ~ f_{\mathbb C}(u^{r}_{\theta}, \sigma_{\theta}); \\
    \tilde{u}^{\kappa}_{\theta, \tau} &= \mu_{\theta} + \sigma_{\theta} \cdot \mathbb{C}
\end{aligned}
\end{equation}
where $f_{\mu_{\theta}}, f_{\sigma_{\theta}}, f_{\mathbb S}$ are fully connected layers. $\mu_{\theta}$ represented the mean of human-robot latent cooperative intents $\mathbf{P}(\tilde{u}^{\kappa}_{\theta, \tau})$, while $\sigma_{\theta}$ denotes the standard deviation of the cooperation range $\mathbf{P}(\tilde{u}^{\kappa}_{\theta, \tau})$. $\mathbb C$ is the sampled instance of cooperation, and $\tilde{u}^{\kappa}_{\theta, \tau}$ refers to the reparameterized policies over total $\tau \in [1, \mathcal{T}]$ .

Additionally, a large-language model (LLM) \cite{gpt} is used as human-robot interface to interpret human language commands. These commands are then converted into robot destinations and human preference configurations (e.g., preferred social distances and social norms) in $I_C$. 

Eventually, NaviDIFF denoises the cooperation distribution during the reverse process of diffusion $p_{\theta}(\cdot|\cdot)$, over $\omega \in [\kappa, 1]$, from $\hat{u}^{\omega}_{\theta, \tau}$ to $\hat{u}^{0}_{\theta, \tau}$ as follows:
\begin{equation}
\begin{aligned}
    \varepsilon_{\theta}^{\omega-1} &= f_{\varepsilon}(\hat{u}^{w}_{\theta,\tau},~ \omega ~|~ I_C)\\
    \hat{u}^{\omega - 1}_{\theta, \tau} & = \frac{1}{\sqrt{1-\alpha_{\omega-1}}}(\hat{u}^{\omega}_{\theta, \tau} - \frac{\alpha_{\omega-1}}{\sqrt{1 - \textstyle{\prod_{k=1}^{\omega-1}}(1-\alpha_k)} } \varepsilon_{\theta}^{\omega-1}) \\ & + \sqrt{\alpha_{\omega-1}} ~ \eta
\end{aligned}
\end{equation}
\noindent where the diffusion is guided by the condition $I_C$, which includes static obstacle information and human preference configuration. Additionally, $\eta \sim \mathcal{N}(\eta; 0, \mathbf{I})$ represents  Gaussian noise, and $\varepsilon_{\theta}$ denotes the conditional denoising term.

After processing through the diffusion neural network, the final policy is sampled from $\mathcal{T}$ policies as $\hat{u}^{r}_{\theta} \sim \{ \hat{u}^{0}_{\theta, \tau}~|~ \tau \in [1, \mathcal{T}] \}$ and is parameterized to connect with the DRL actor. Notably, the gap between the continuous PH physical process and the discrete DRL decision-making process is addressed by Eq.~(\ref{equation:gap}).

\begin{figure*}[!t]
\centering
\includegraphics[width=0.98\linewidth]{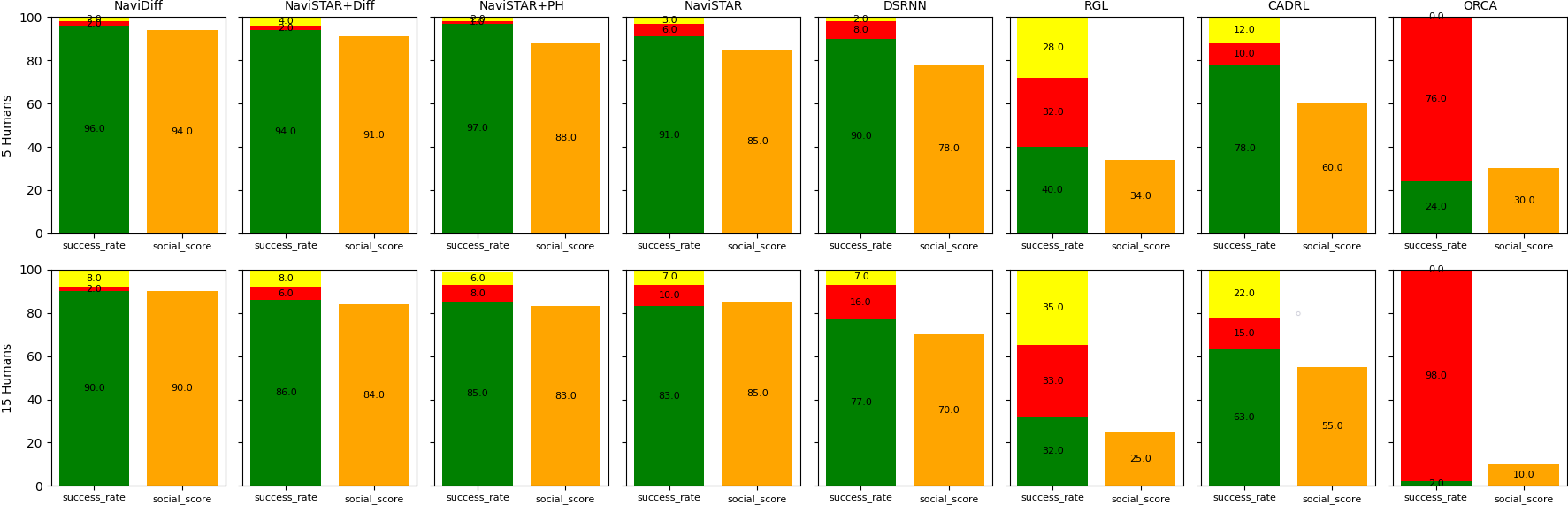}
\vspace{-3pt}
\caption{\small Experiment Results: Success rate (green), collision rate (red), timeout rate (yellow), and social score (orange) for each policy, based on 500 test runs under the same environment configuration.}
\vspace{-10pt}
\label{fig:datatable}
\end{figure*}

\subsection{Training Procedure}

The training procedure for NaviDIFF consists of the following key stages: 1) Proximal policy optimization (PPO)-based policy initialization \cite{dsrnn1, ppo}; 2) RLHF reward learning \cite{wang2023navistar,lee2021bpref}; and 3) off-policy RL policy learning \cite{wang2023navistar, haarnoja2018soft}. Firstly, we employ a state-of-the-art on-policy RL algorithm, PPO \cite{ppo}, to pre-train an initial social robot policy $\pi_{0}$, optimizing the objective functions $\mathcal{L}_{\theta}$ and $\mathcal{L}_{\phi}$ in Eqs.~(\ref{PPO-actor}) and (\ref{PPO-critic}). PPO updates policy gradients based on action advantages, $(A_t = A_{\pi}(a_t, s_t) = Q_{\phi}(a_t, s_t) - V_{\pi}(s_t) )$, and minimizes a clipped-ratio loss over mini batches from the recent dataset (generated by $\pi$), with clip rate $\epsilon$, as follows:
\begin{equation}\label{PPO-actor}
\begin{aligned}
\mathcal{L}_{\theta} = & - \mathbb{E}_{\tau \sim \pi_{\theta}} [ min(\frac{\pi_{\theta}(a_t|s_t)}{\pi_{\theta_{old}}(a_t|s_t)}A_t ~,~ \\ 
& clip( \frac{\pi_{\theta}(a_t|s_t)}{\pi_{\theta_{old}}(a_t|s_t)}, 1 \pm \epsilon)A_t  ) ].    
\end{aligned}
\end{equation}
The state-value estimator, $V_{\phi}(s_t)$ is learned by the critic and regressed against the target of discounted returns using Generalized Advantage Estimation \cite{gae} as follows:
\begin{equation}\label{PPO-critic}
\mathcal{L}_{\phi} = \mathbb{E}_{\tau \sim \pi_{\theta}} \left[ ( V_{\phi}(s_t) - V_{\hat{\phi}}(s_t) )^{2} \right].
\end{equation}
Once agent policy is initialized by pre-train stage, we embed human preferences and expectations on policy learning via RLHF fine-tuning procedure from \cite{wang2023navistar}.



\section{Experiments And Results}
\subsection{Simulation Experiment Setup}
\noindent\textbf{Environment Configuration: }We conduct the simulation experiments using the OpenAI Gym-based social robot navigation environment \cite{wang2023navistar}, with Hamiltonian-based environmental dynamics as defined in Eq.~(\ref{equation:gap}). Pedestrians are controlled by ORCA \cite{orca} or Social Force (SF) \cite{sf} personal policies, with the assumption that the mobile robot is invisible to pedestrians to ensure maximum social security training. 

At each timestep $t$ of training procedure, the robot updates its local observation $({o^r} = {s^r, s^{ho}})$ and generates an action $a^r = [v_x, v_y]$ based on the policy function $\pi_{\theta}$, aiming to maximize the objective $\mathcal{J} = \arg\max \mathbb E [\sum \gamma^{\mathrm{t}} \mathcal{R}(s_t, a_t)]$. The robot's self-state is defined as $s^r = [ p_{ x}, p_{ y}, v_{{x}}, v_{{y}}, \rho , g_x, g_y, v_{pref}] $, where $p_x, p_y$ represent position, $v_{{x}}, v_{{y}}$ represent velocity, $\rho$ is the robot's  personal radius, $g_x, g_y$ are goal positions, and $v_{pref}$ is the preferred velocity. 

Human states are divided into public and private components: the public state $s^{ho}$ includes $[p_x, p_y, v_x, v_y, \rho]$, while the private state $s^{hu}$ includes $[g_x, g_y, v_{pref}]$. Static obstacles are represented by motionless pedestrians, who share the same state structure as the moving pedestrians. Finally, the simulation is carried out in an open space with dimensions of ${20~m\times20~m}$.


\begin{table}{}{}
    \centering 
\centering
\resizebox{.45\textwidth}{!}{
\begin{tabular}{cccccccc}
\toprule
& \multicolumn{3}{c}{ Success Rate } & & \multicolumn{3}{c}{ Social Score } \\
\cline { 2 - 4 } \cline { 6 - 8 } Methods & \multicolumn{3}{c}{ Human Number } & & \multicolumn{3}{c}{ Human Number } \\
& 5 & 10 & 15 & & 5 & 10 & 15 \\
\midrule 
ORCA \cite{orca}& $24$ & $9$ & $2$ & & $30$ & $13$ & $10$ \\
SF \cite{sf}& $40$ & $22$ & $8$ & & $35$ & $25$ & $14$ \\
CADRL \cite{CADRL}& $78$ & $70$ & $63$ & & $60$ & $58$ & $55$ \\
SARL \cite{sarl}& $84$ & $77$ & $65$ & & $65$ & $61$ & $58$ \\
RGL \cite{rgl}& $40$ & $36$ & $32$ & & $34$ & $30$ & $25$ \\
DSRNN \cite{dsrnn1}& $90$ & $83$ & $77$ & & $78$ & $75$ & $70$ \\
NaviSTAR \cite{wang2023navistar} & $91$ & $90$ & $83$ & & $85$ & $82$ & $85$\\
NaviSTAR+PH & $\textbf{97}$ & $85$ & $85$ & & $88$ & $85$ & $83$\\
NaviSTAR+Diff & $94$ & $88$ & $86$ & & $91$ & $90$ & $84$\\
NaviDIFF   & $96$ & $\textbf{92}$ & $\textbf{90}$ & & $\textbf{94}$ & $\textbf{93}$ & $\textbf{90}$\\ 
\bottomrule
\end{tabular}}
\caption{\small Experiment Results: Success rate and social score for the proposed method, baselines, and ablation models across different numbers of humans. Bold numbers indicate the best performance among the tested methods.}
\vspace{-20pt}
\label{table:exp1}
\end{table}

\noindent\textbf{Baselines and Ablation Models: }We selected a set of current state-of-the-art social navigation planners as baselines for testing, including: ORCA \cite{orca} and SF \cite{sf} as conventional methods; CADRL \cite{CADRL} and SARL \cite{sarl} as MLP-based and deep V-learning approaches; RGL \cite{rgl} representing a graph convolution network and model-based RL baseline; DSRNN \cite{dsrnn1}, an RNN-based approach; and NaviSTAR \cite{wang2023navistar}, which represents transformer-based planners.

Additionally, we include two ablation models for comparison: NaviSTAR+PH, which combines the NaviSTAR framework with the PH dynamic framework, and NaviSTAR+Diff, which incorporates the diffusion network into NaviSTAR to evaluate the performance of each component of NaviDIFF.

All baseline algorithms were trained according to their original papers and configurations. Subsequently, we tested all policies in the same environment under identical conditions. For each ablation model, we used a learning rate of $4 \times 10^{-5}$ and trained for $1 \times 10^4$ episodes.


\noindent\textbf{Evaluation Metrics: }As shown in Fig.~\ref{fig:datatable}, two evaluation metrics are used in the simulation experiments. The success rate (SR) is calculated as the number of successful navigation cases out of a total of 500 testing cases. The social score (SS) \cite{wang2023navistar} evaluates the overall navigation performance, considering both social norm compliance and path quality.


\subsection{Quantitative Analysis}
\begin{figure}
\centering
\vspace{-2pt}
\subfloat[NaviSTAR+PH]{\label{fig:a} \includegraphics[width=4.0CM]{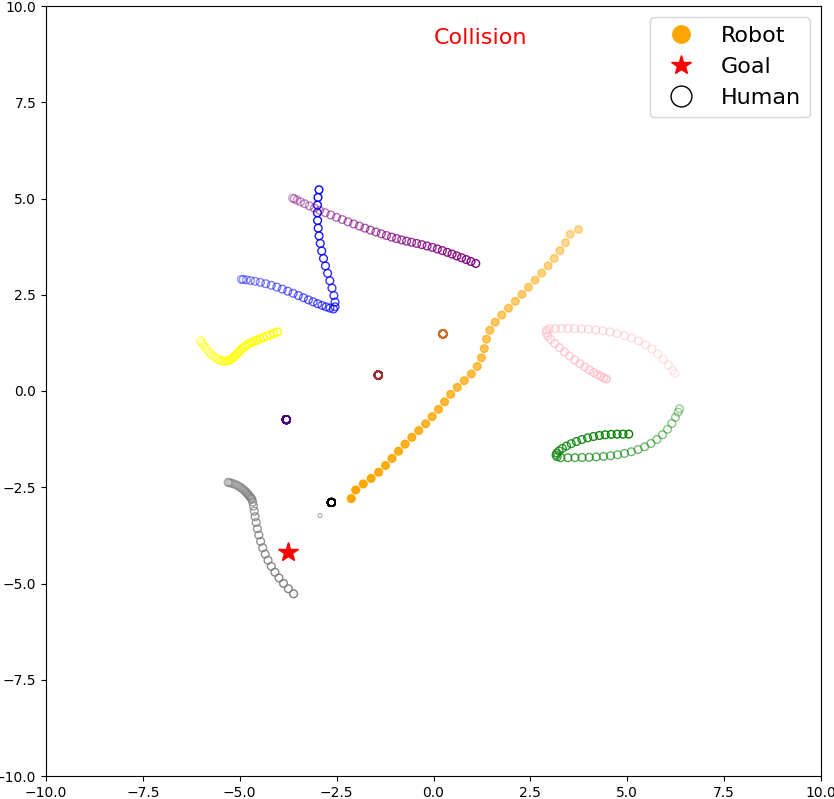}}~~
\subfloat[NaviDIFF]{\label{fig:b}\includegraphics[width=4.0CM]{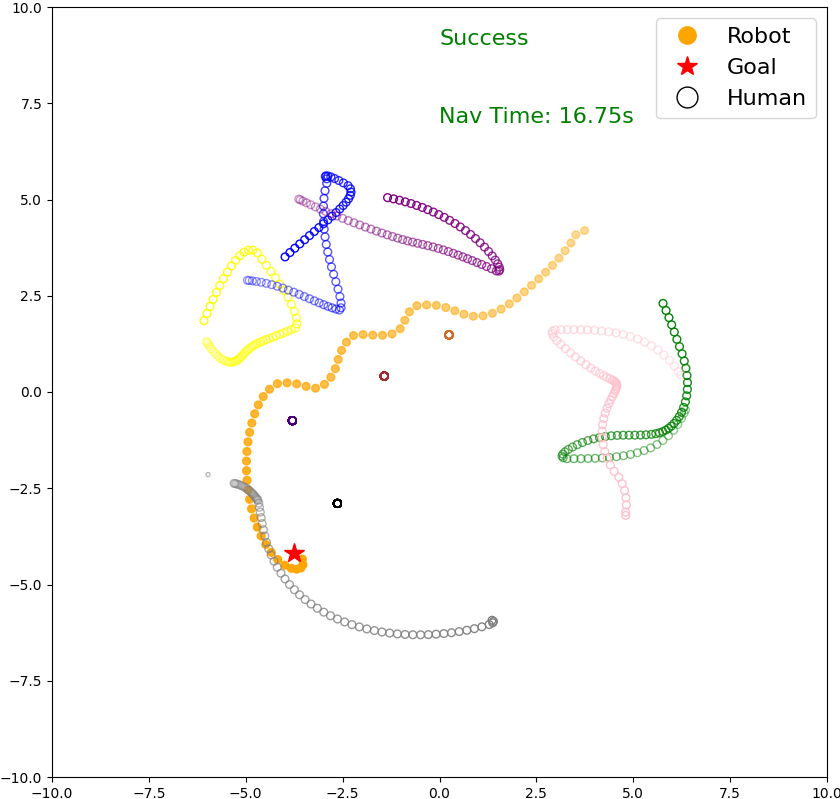}}
\vspace{-1pt}
\hspace{1pt}
\subfloat[NaviSTAR+Diff]{\label{fig:e}\includegraphics[width=4.0CM]{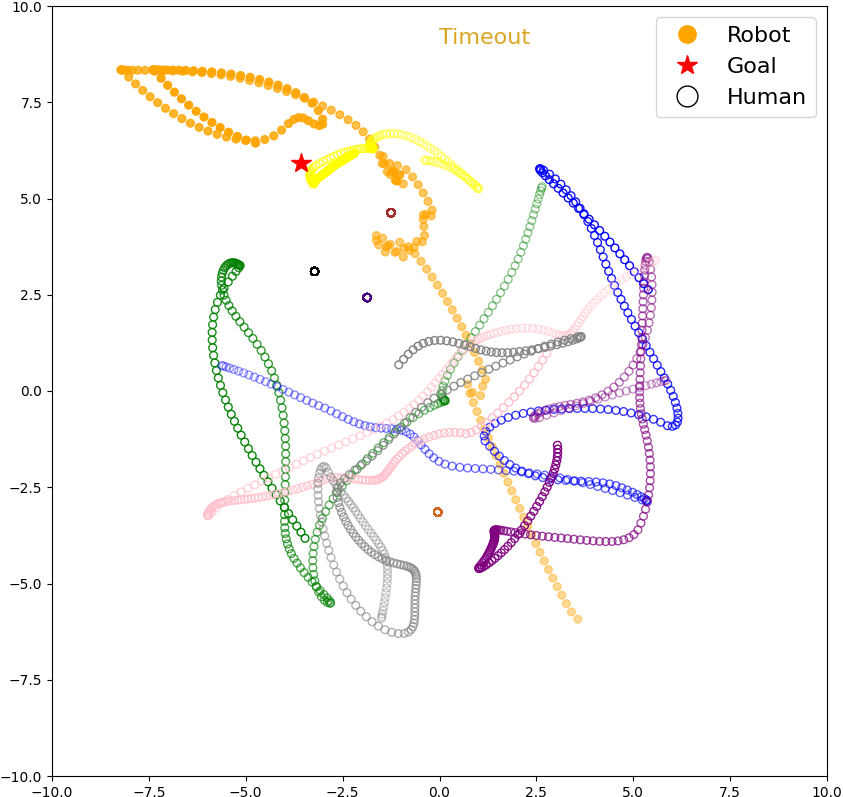}}~~~
\subfloat[NaviDIFF]{\label{fig:f}\includegraphics[width=4.0CM]{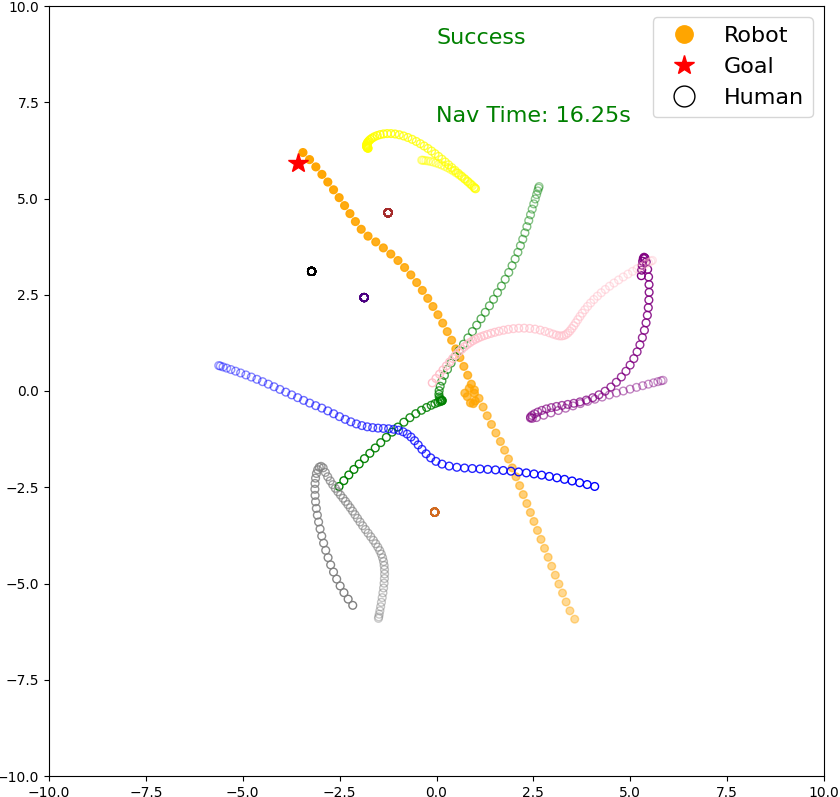}}
\vspace{-0pt}
\caption{\small Comparison of Trajectories Visualization: Visualization of the trajectories for NaviDIFF and two ablation models, tested on two different scenarios. Zoom in for better readability.}
\vspace{-10pt}
\label{fig:TRAJECTORY}
\end{figure}
As shown in Fig.~\ref{fig:datatable} and Table~\ref{table:exp1}, conventional planner ORCA and SF struggled in dynamic crowd environments, particuarly under the robot's invisible configuration. Their lowest SR and SC scores indicate the limitations of short-sighted reactive planners, which only demonstrate avoidance behaviors just before imminent collisions.

In Fig.~\ref{fig:datatable} and Table~\ref{table:exp1}, previous learning-based planners, such as CADRL, SARL, RGL, and SRNN, exhibited limited navigation abilities and weak social norm compliance due to their restricted HRI dependency representation, relying on fully connected layers, convolutional networks, or recurrent networks. Specifically, RGL's model-based RL execution is highly parameter-sensitive, making it difficult to adapt to different environmental configurations. While SRNN uses an ST-graph and recurrent framework to capture spatial-temporal correlations between agents, its handcrafted reward function makes it challenging to achieve satisfactory social norm compliance.

NaviSTAR demonstrated better performance compared to the other baselines; however, NaviDIFF outperforms NaviSTAR, especially as the number of humans increases. This improvement is attributed to NaviDIFF's energy-based interactive feature descriptions, which enhance HRI inference.

As shown in Fig.\ref{fig:TRAJECTORY}, NaviDIFF significantly outperforms the ablation models in both collision avoidance and social norm compliance. Additionally, NaviDIFF demonstrates more flexible HRI inference and better environmental energy balance representation compared to NaviSTAR+Diff, as seen in Fig.~\ref{fig:f} and Fig.~\ref{fig:e}. Notably, in Fig.~\ref{fig:b}, NaviDIFF adjusts its orientation to maintain social distance when pedestrians approach, while the ablation model NaviSTAR+PH leads to a collision, misinterpreting the static pedestrian’s cooperative distribution. These findings suggest that incorporating both a PH representation network and a diffusion network can greatly enhance the performance and social compliance of socially-aware navigation algorithms in human-filled environments.

\subsection{Real-world Robot Demonstration}
\begin{figure}[h]
\centering
\vspace{-5pt}
\includegraphics[width=0.9\linewidth]{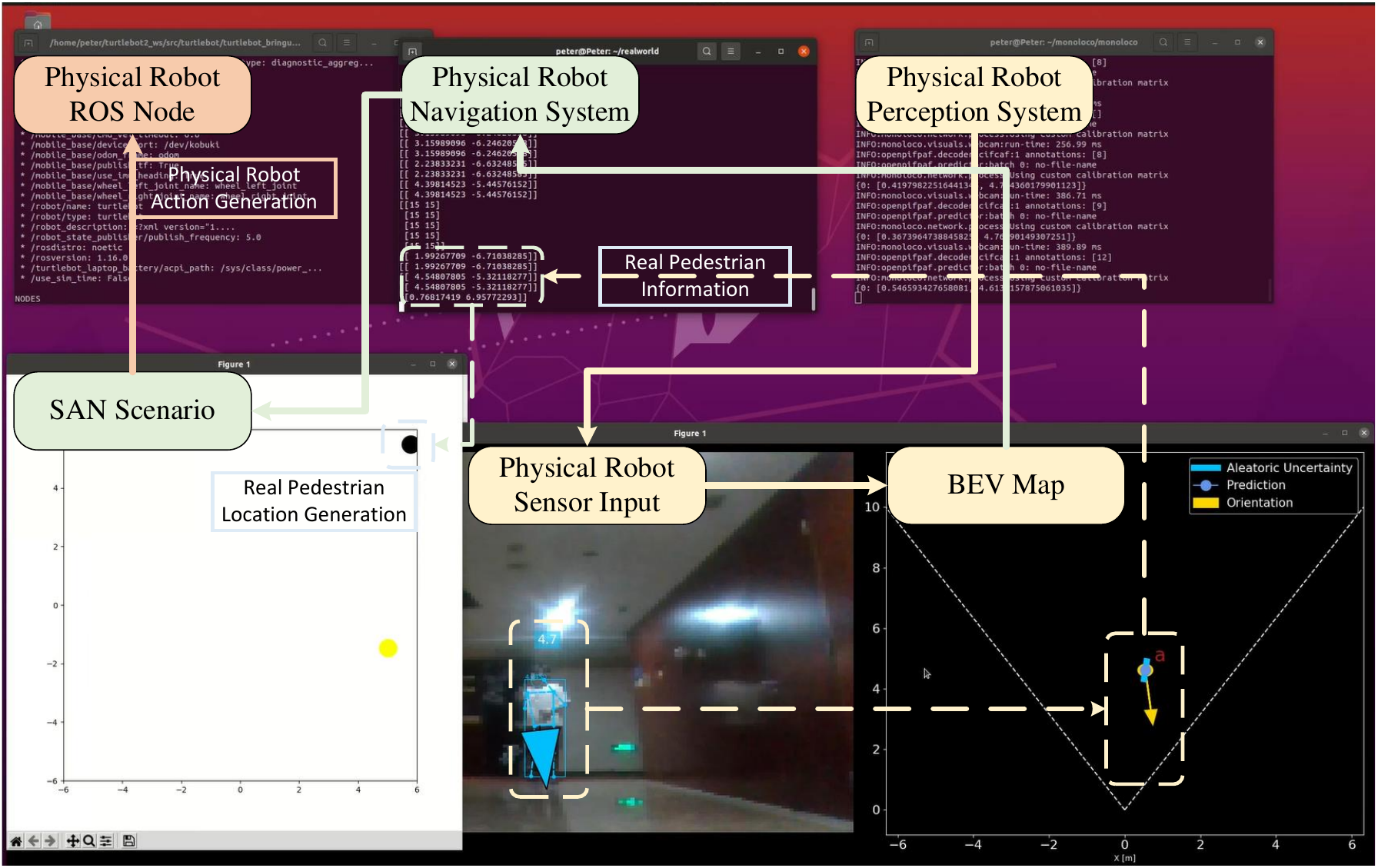}
\vspace{-5pt}
\caption{\small Configuration of the physical mobile robot platform.}
\vspace{-10pt}
\label{fig:physical}
\end{figure}

We also tested the proposed navigation method on a real robot platform. As depicted in Fig.~\ref{fig:physical}, the physical mobile robot consists of a robot perception system, a social navigation system node, and a physical robot ROS node. Pedestrian information is captured by the robot’s sensor (in our experiments, the Realsense D435 camera), which converts raw RGB data into a bird’s-eye view map \cite{bertoni2019monoloco, vats2023enhancing}. This pedestrian location data is then processed by the social navigation system node, where the pre-trained NaviDIFF model is used to initialize the SAN scenario and generate robot actions. The robot is then controlled using the NaviDIFF model via the physical robot ROS node.

Demonstration videos showcasing the real-world performance of the robot can be found in the supplementary material.


\vspace{-6pt}
\section{Conclusion}
\vspace{-5pt}
In this paper, we propose NaviDIFF, a diffusion-based, physics-informed social navigation planner for addressing the HCSAN problem. By leveraging a transformer network to capture social and temporal HRI features and exploring latent human-robot cooperative uncertainty via a diffusion model, NaviDIFF achieves superior performance compared to previous methods.

For future work, we plan to extend NaviDIFF to handle more complex environments, explore multi-robot interactions, and improve computational efficiency for real-time applications.


\newpage
\typeout{}
\bibliography{main}
\bibliographystyle{IEEEtran}
\end{document}